\documentclass{article}

\usepackage[nonatbib, final]{neurips_2025}

\usepackage[utf8]{inputenc} % allow utf-8 input
\usepackage[T1]{fontenc}    % use 8-bit T1 fonts
\usepackage{hyperref}       % hyperlinks
\usepackage{url}            % simple URL typesetting
\usepackage{booktabs}       % professional-quality tables
\usepackage{amsfonts}       % blackboard math symbols
\usepackage{nicefrac}       % compact symbols for 1/2, etc.
\usepackage{microtype}      % microtypography
\usepackage{xcolor}         % colors

\usepackage{amsmath}

\usepackage{amsthm}

\theoremstyle{definition}

\usepackage{tabularx}
\usepackage{multicol}
\usepackage{multirow}
\usepackage{rotating}
\usepackage{diagbox}
\usepackage{graphicx}
\usepackage{tikz}
\usetikzlibrary{positioning, shapes.geometric, fit}

\newcommand{\lratio}[1]{\setlength{\hsize}{#1\hsize}} 
\newcolumntype{Y}{>{\lratio{0.65}}>{\centering\arraybackslash}X}

\title{Addressing the Cold-Start Problem for Personalized Combination Drug Screening}

\author{%
  Antoine de Mathelin, Christopher Tosh and Wesley Tansey \\
  Computational Oncology\\
  Memorial Sloan Kettering Cancer Center\\
  New York, NY 10065 \\
  \texttt{dematha@mskcc.org} \\
}

\begin{document}

\maketitle

\begin{abstract}
Personalizing combination therapies in oncology requires navigating an immense space of possible drug and dose combinations, a task that remains largely infeasible through exhaustive experimentation. Recent developments in patient-derived models have enabled high-throughput ex vivo screening, but the number of feasible experiments is limited. Further, a tight therapeutic window makes gathering molecular profiling information (e.g. RNA-seq) impractical as a means of guiding drug response prediction. This leads to a challenging cold-start problem: how do we select the most informative combinations to test early, when no prior information about the patient is available? We propose a strategy that leverages a pretrained deep learning model built on historical drug response data. The model provides both embeddings for drug combinations and dose-level importance scores, enabling a principled selection of initial experiments. We combine clustering of drug embeddings to ensure functional diversity with a dose-weighting mechanism that prioritizes doses based on their historical informativeness. Retrospective simulations on large-scale drug combination datasets show that our method substantially improves initial screening efficiency compared to baselines, offering a viable path for more effective early-phase decision-making in personalized combination drug screens.
\end{abstract}

\section{Introduction}

Identifying effective drug combinations for individual patients remains one of the central challenges in precision oncology \cite{al2012combinatorialTherapy, o2016unbiased}. Recent advances in patient-derived models (e.g. spheroids, tumoroids, and organoids) allow the \textit{ex vivo} testing of drug combinations at multiple doses, providing a promising platform for tailoring treatments to individual patients \cite{sachs2018living, vlachogiannis2018patient, wensink2021patient}. However, even with these tools, the problem is challenging: the combinatorial space of drug combinations and dose levels is vast \cite{tosh2025batchie}. For instance, a full pairwise drug screen of 100 drugs and at 100 dose-levels would require $\simeq 50$M experiments. Exhaustively screening all combinations is thus practically intractable.

% For instance, the Bayesian Active Treatment Combination Hunting via Iterative Experimentation (BATCHIE) \cite{tosh2025batchie} framework 

% The main active learning approaches consists of selecting a batch of experiments based on an optimal coverage of the design space or based on the prediction uncertainty of a trained machine learning model \cite{settles2009active}. This implies that a feature representations (embeddings) of the drug is available 

% To address this challenge, experimental design and active learning approaches aim to optimize the batch of experiments by selecting drug-dose combinations that are most informative \cite{yu2021currentALdrugdisco, bailey2023deepbatchALdrug, reker2015activeLdrugdisco, tosh2025batchie}

Active learning frameworks have been developed that aim to efficiently explore the combination landscape through iterative batches of experiments~\cite{bertin:etal:2023:recover,tosh2025batchie}. These frameworks seek to select drug-dose combinations that are most informative based on the results of previous batches. They use prior drug response data across a cohort of cell lines to learn a model of drug responses, capturing functional similarity between treatments. Likewise, a predictive model of patient response — summarizing how each sample responds to a range of drugs — is learned based on the prior drug responses for the sample and any available molecular profiling information (e.g. somatic alterations or transcriptomics).

Challenges arise when translating drug screens from a preclinical tool on cell lines to a clinical diagnostic on ex vivo samples. First, molecular profiling is often too slow, too inconsistent, or too expensive to generate within the therapeutic window for each new patient~\cite{byron:etal:2016:translating-rna-clinic}. Second, unlike cell lines which can be propagated indefinitely, ex vivo screens seek to minimize passaging of samples and thus are limited to a relatively small number of experiments per patient, often as few as two iterative batches~\cite{peterziel:etal:2022:inform-trial}.
%Current precision oncology clinical trials have thus opted for simply pre-designing a panel of single agents, potentially combining them based on expert knowledge~\cite{peterziel:etal:2022:inform-trial}, despite expert recommendations often not having ideal cytotoxicity empirically~\cite{tosh2025batchie}.
Scaling precision oncology ex vivo screens to combinations thus leads to a central question: \textbf{Given precious few samples and a lack of molecular profiling, how do we choose informative drugs for the first batch of experiments? }

%By comparison, emerging clinical trials can screen hundreds of compounds against patient samples in less than 7 days for a modest price~\cite{peterziel:etal:2022:inform-trial}. However, 

% A major limitation arises at the onset of treatment for a new patient: we lack any drug response data and thus cannot construct a patient representation.
We refer to this as a \textbf{cold-start problem}, analogous to those faced in recommendation systems when a new user enters with no interaction history \cite{lam2008addressing, schein2002methods}. %We assume no access to transcriptomic or other biological omics data at this stage, either due to time constraints or feasibility — a realistic scenario in many clinical settings. As a result, we are unable to initialize a biologically meaningful representation of the patient.
% Our work focuses on this early-phase, high-uncertainty setting. We ask: \textbf{How should we choose the very first batch of drug-dose combinations to test on a new patient, in the absence of any prior drug response or biological profile?}
In such a scenario, random selection is often seen as a reasonable default. Indeed, if the new patient exhibits biological behavior completely different from any previously seen patients, no drug-dose selection strategy would outperform random choice. However, we assume that one can exploit prior knowledge about drug and dose relationships, derived from historical drug response data on other patients, to guide a more informative first set of experiments for the new patient.

We develop a framework that leverages a pretrained deep learning model fitted on a cohort of previous patients to design the initial drug-dose batch for new patients. The deep learning model is trained to predict the viability of the tumor against a drug combination for any dose pair. The model is then used to provide a unique representation (embedding) for each drug combination as well as an importance score for each dose. Our approach to select the first batch then combines:
\begin{itemize}
\item $K$-medoids clustering on drug combination embeddings to identify a diverse and representative subset of drug combinations.
\item Dose selection based on a reweighting scheme provided by the importance scores.
\end{itemize}

We validate the effectiveness of our approach through retrospective simulation on large-scale drug combination datasets. We simulate the cold-start scenario by initializing experiments without patient-specific data. Our method consistently outperforms baseline selection strategies, demonstrating its ability to prioritize informative drug-dose pairs even in the absence of prior biological or response information.

\section{Related Work}

% Deep learning has been widely applied to drug response prediction tasks in recent years, with models leveraging chemical structure, gene expression, and prior pharmacogenomic data to forecast treatment outcomes [1,2,3]. Many of these approaches focus on monotherapy prediction or cell line panels with known omics features. In our case, we do not assume access to such biological features for the patient at prediction time. Instead, our model relies on learned representations of drugs, doses, and patients, trained end-to-end from historical drug response measurements.

% [1] Kuenzi et al., Predicting drug response and synergy using deep learning, Cell Systems, 2020
% [2] Manica et al., Toward explainable drug sensitivity prediction via multimodal attention-based convolutional encoders, Bioinformatics, 2019
% [3] Sakellaropoulos et al., A deep learning framework for predicting response to therapy in cancer, Cell Reports, 2019

\subsection{Deep Learning for Drug Response Prediction}

Our work is grounded in deep learning approaches for drug response prediction. In recent years, deep learning has been widely applied to drug response prediction tasks, with models leveraging chemical structure, gene expression, and prior pharmacogenomic data to forecast treatment outcomes \cite{kuenzi2020predictingDLDR, manica2019towardDLDR, sakellaropoulos2019deepDLDR}. Recent deep transfer learning strategies have further improved generalization by leveraging knowledge from related tasks or datasets \cite{chen2022deepTLDR, mourragui2021predictingTLDR, sederman2024screendlTLDR}. These approaches adapt drug response models trained on cell lines to patient-derived models of cancer via transfer learning. While these methods have shown promising results, they typically assume the availability of rich biological features (e.g., RNA-seq), which may not be available in early clinical or organoid testing scenarios.

% Many of these approaches focus on monotherapy prediction using large-scale cell line datasets, often integrating omics features such as gene expression to personalize predictions.

Our approach differs in that we do not assume access to omics data at test time. Instead, we construct patient and drug representations directly from prior drug response data and rely on active experimentation to gather informative data for new patients, following the BATCHIE framework \cite{tosh2025batchie}.

\subsection{Active Learning and Experimental Design for Drug Discovery}

% Active learning has gained increasing attention for drug discovery and experimental design, as it promises to reduce experimental cost by selectively querying informative data points. Several studies have developed acquisition functions tailored to molecular property prediction, virtual screening, or lead optimization \cite{yu2021currentALdrugdisco, bailey2023deepbatchALdrug, reker2015activeLdrugdisco, jimenez2020drugdiscoBA, graff2021acceleratingdrugdiscoBA, yang2021efficientdrugdiscoBA}.

Active learning (AL) is a class of machine learning techniques aimed at selecting the most informative data points to label, thereby reducing the total number of required annotations while maintaining high predictive performance \cite{settles2009active}. In its standard setting, AL assumes access to a large pool of unlabeled data and a limited labeling budget. The goal is to iteratively query samples that are expected to improve model performance the most, typically based on uncertainty sampling, diversity sampling, or expected model change \cite{deheeger2021discrepancy, gal2017deep, lewis1995sequential}. Active learning has gained increasing attention for drug discovery and experimental design, as it promises to reduce experimental cost by selectively querying informative data points. Several studies have developed acquisition functions tailored to molecular property prediction, virtual screening, or lead optimization \cite{ bailey2023deepbatchALdrug,
graff2021acceleratingdrugdiscoBA,
jimenez2020drugdiscoBA, 
reker2015activeLdrugdisco, yang2021efficientdrugdiscoBA, yu2021currentALdrugdisco}. However, these approaches often rely on having initial features or representations of the input samples — for example, gene expression profiles or drug response vectors — which are not available in our cold-start setting. This makes standard AL techniques difficult to apply, as they cannot operate without some prior information about the new patient.

% In our work, we frame the problem as a two-stage experimental design: selecting representative drug pairs and then informative dose points per pair. We build on methods from optimal experimental design [14,15], but apply them in a deep learning setting with learned representations and surrogate models.

% Notably, most prior work on active learning assumes a warm-start scenario (e.g., access to biological features or some prior responses). 

In this work, we explore the cold-start active learning problem for combinatorial drug screening, where we must select an initial batch of experiments without any patient-specific biological information. Our approach leverages prior drug response data from other patients to construct representations of drug combinations. We then apply the $K$-medoids clustering algorithm \cite{rdusseeun1987clustering} to select a diverse and representative subset of drug combinations for the initial screen.

\subsection{Cold-Start in Drug Response and Combinatorial Settings}

The cold-start problem — where no prior data is available for a new patient or condition — presents a significant challenge for drug response prediction. In recommendation systems, this issue is commonly addressed through meta-learning strategies \cite{lee2019melu, vartak2017meta}. In the context of drug combination screening, Chen et al. \cite{chen2024makingyourfirstchoice} and Dewulf et al. \cite{dewulf2021coldstartdrugdrug} have investigated cold-start scenarios where an initial batch of experiments must be selected before model training. These approaches typically assume access to a fixed feature representation of the new instance. In contrast, our setting is more constrained: while we leverage a pretrained model, we assume no features or biological information are available for the new patient, making the techniques from prior work inapplicable to our problem.

\section{Cold-Start Drug-Dose Combination Selection}

In this section, we describe our approach to selecting an informative initial batch of drug-dose experiments in the cold-start setting. Our method builds on a pretrained drug response model and leverages its learned representations to guide batch selection. The section is structured as follows: first, we present the drug response model used to learn embeddings and dose importance scores from historical data; then, we detail the selection strategy that uses these outputs to construct a diverse and informative initial experimental batch for a new patient.

\subsection{Drug Response Model}

% Deep learning, how we enforce monotonicity constraints, how we model drug combination, sum of rank1 matrices. Drug Embedding, Patient embedding, experimental condition embedding.

% We model the drug response function using a deep learning framework that maps the sample ID, drugs IDs, to a full drug response curve expressing the viability of the sample against this drug combination at different doses. Specifically we divided the log-dose-level in 100 bins. Thus, for pairwise combinations of drug, the model takes two drug ID plus the sample ID and produce a matrix R 100 times 100 where R{ij} is the predicted viability for dose i and j.

% information to a predicted viability or therapeutic effect. This function serves as the core predictive engine throughout our experimental design strategy, and is trained using historical response data across a large number of patients and drug combinations.

We model the drug response function using a deep learning framework that predicts the viability of a biological sample in response to pairwise drug combinations across a range of doses. Let $s \in \mathcal{S}$ denote a sample (e.g., a patient sample or cell line), and let $d_1, d_2 \in \mathcal{D}$ denote two drugs from a set of available drugs $\mathcal{D}$. We discretize the log-transformed dose space into $L = 100$ bins, representing different dose levels. The deep learning model learns a function
\[
f_\theta: \mathcal{S} \times \mathcal{D} \times \mathcal{D} \rightarrow \mathbb{R}^{L \times L}
\]
parameterized by $\theta$, which maps a sample $s$ and a drug pair $(d_1, d_2)$ to a response matrix $R^{(s, d_1, d_2)} \in \mathbb{R}^{L \times L}$, where each entry $R^{(s, d_1, d_2)}_{i,j}$ represents the predicted viability of sample $s$ at dose level $i \in [1, 100]$ of drug $d_1$ and dose level $j \in [1, 100]$ of drug $d_2$. This formulation captures the full dose-response surface for each drug combination and enables downstream selection strategies.

\subsubsection{Model Architecture}

% Our model takes as input a triplet: the patient index, the two drugs under test (drug 1 and drug 2), and their associated dose levels. These inputs are passed through respective embedding layers:
% \begin{itemize}
% \item Patient embeddings encode latent patient-specific drug sensitivity,
% \item Drug embeddings capture pharmacological similarity derived from response data,
% \item Experimental condition embeddings represent discrete dose levels or normalized dose scalars.
% \end{itemize}

% Our model $f_\theta$ takes as input a triplet consisting of a sample (or patient) index $s \in \mathcal{S}$ and two drug indices $d_1, d_2 \in \mathcal{D}$. Each of these inputs is passed through a corresponding embedding layer to obtain fixed-length latent vectors in $\mathbb{R}^p$:
% \begin{itemize}
% \item The sample embedding $e_s \in \mathbb{R}^p$ corresponding to $s$,
% \item The drug embeddings $e_{d_1}, e_{d_2} \in \mathbb{R}^p$ corresponding to $d_1, d_2$,
% \end{itemize}
% These embeddings are concatenated into a single vector in $\mathbb{R}^{5p}$ and passed through a multi-layer perceptron (MLP) to predict the drug response value. 

Our model $f_\theta$ takes as input a triplet consisting of a sample index $s \in \mathcal{S}$ and two drug indices $d_1, d_2 \in \mathcal{D}$. Each input is passed through a corresponding embedding layer to obtain fixed-length latent vectors in $\mathbb{R}^p$:
\begin{itemize}
\item The sample embedding $e_s \in \mathbb{R}^p$ corresponding to $s$,
\item The drug embeddings $e_{d_1}, e_{d_2} \in \mathbb{R}^p$ corresponding to $d_1$ and $d_2$.
\end{itemize}

We then construct two input vectors by concatenating the sample embedding with each drug embedding: $[e_s; e_{d_1}] \in \mathbb{R}^{2p}$ and $[e_s; e_{d_2}] \in \mathbb{R}^{2p}$. These concatenated vectors are independently passed through a shared multi-layer perceptron (MLP), producing two dose-dependent latent matrices:
\[
M(s, d_1) \in \mathbb{R}^{L \times r}, \quad M(s, d_2) \in \mathbb{R}^{L \times r},
\]
where $r \in \mathbb{N}$ is a tunable hyperparameter and $L$ represents the number of discretized log-dose levels.

% The final predicted drug response surface $R^{(s, d_1, d_2)} \in \mathbb{R}^{L \times L}$ is obtained by taking the matrix product of these two representations:
% \[
% R^{(s, d_1, d_2)} = M(s, d_1) \cdot M(s, d_2)^\top.
% \]

This modeling approach allows us to decompose the prediction task into two distinct components: sample–drug interaction and drug–drug interaction. The MLP processes the concatenation of the sample embedding with each drug embedding individually to produce two matrices  $M(s, d_1), M(s, d_2) \in \mathbb{R}^{L \times r}$ , which capture how each drug affects the sample across dose levels. The final drug response matrix is computed via matrix multiplication:
\[
R^{(s, d_1, d_2)} = M(s, d_1) \cdot M(s, d_2)^\top,
\]
enabling the model to express interactions between drugs through the structure of the resulting matrix. The rank  $r$  controls the degree of interaction: when  $r = 1$ , the response reduces to the outer product of two dose-wise response vectors, effectively modeling the combination effect as a multiplicative function of the individual drug responses. As  $r$  increases, the model can capture richer, higher-order interactions.

This factorized structure is motivated by two practical considerations. First, it facilitates the enforcement of monotonicity constraints on the dose-response curve, which we describe in the next subsection. Second, choosing  $r < L$  significantly reduces the number of parameters, enhancing scalability. This is particularly important since we ultimately aim for a methodology that generalizes to more complex combinations (e.g., 3 or more drugs), where parameter efficiency becomes critical.

\subsubsection{Monotonicity and Smoothness Constraints}

% To ensure biologically realistic behavior, we enforce monotonicity with respect to dose — increasing the dose of a drug should not result in increased cell viability (i.e., a worse response), except in rare edge cases. 

To ensure biologically realistic behavior, we enforce monotonicity with respect to dose. As all drugs in our library are cytotoxic agents, we assume that increasing the dose of a drug will not result in increased cell viability in expectation. To implement this constraint, our MLP does not directly predict the response matrix \( M(s, d) \in \mathbb{R}^{L \times r} \). Instead, it outputs a matrix of "dose-wise derivatives" \( \Delta M(s, d) \in \mathbb{R}^{L \times r} \), which we interpret as incremental changes in viability along increasing dose levels.

We enforce non-negativity of these increments by applying an elementwise absolute value: \( \Delta M := |\Delta M| \). The cumulative sum of these values along the dose axis gives us the preliminary response matrix: $
\widetilde{M}(s, d) = \text{cumsum}(\Delta M(s, d), \text{ axis}=0). $

To ensure that the resulting values remain bounded within \([0, 1]\), we normalize each matrix by the total cumulative sum across the dose dimension: $
M(s, d) = \widetilde{M}(s, d) \big/ \sum_{i=1}^{L} \Delta M_{i, :}$.

Finally, to introduce smoothness into the dose-response profile and avoid spurious step-like behaviors, we apply a 1D convolution with a Gaussian kernel along the dose axis of each column of \( M(s, d) \). This results in a smooth, monotonic matrix that models the drug’s effect across doses in a biologically plausible way.

\subsubsection{Pretraining and fine-tuning}

% The model is trained using mean absolute error (MAE) between predicted and observed viability values, across all available patient–drug–dose combinations. We pretrain the model on a large panel of historical data, allowing it to learn meaningful representations for drugs and doses. In downstream use cases, such as our experimental design setting, the model can then be adapted or used as a fixed oracle to guide decision-making.

The model is trained using the mean absolute error (MAE) between predicted and observed viability values, aggregated over all available sample–drug–dose combinations. We pretrain the model on a large historical dataset comprising multiple patients or samples, allowing it to learn representations (embeddings $e_s, e_d$) for drugs and doses. However, when deploying this pretrained model to a new patient \( s^\ast \), we face a fundamental limitation: there is no embedding \( e_{s^\ast} \in \mathbb{R}^p \) available for this unseen sample. To address this, we must first select an initial batch of drug combinations to test on the new patient, obtain the corresponding viability responses, and use this small dataset to initialize and fine-tune a new embedding for \( s^\ast \). Typically, we initialize \( e_{s^\ast} \) randomly and optimize it (and optionally the rest of the model) to minimize prediction error on the newly collected observations. Figure \ref{fig:drug_response_surface} presents an example of drug response obtained by the pretrained and fine-tuned models. 

\subsection{Cold Start Selection Strategy}

% Setting, we have to choose K drug pairs and dose pairs. Our methodology is the following.

% \begin{itemize}
%     \item Choose the drug pairs
%     \item Choose the dose pairs
% \end{itemize}

% Dose pair. For every drug in the set of drug pair, we have a grid to cover. We can reweight the grid based on the second derivative.

% Drug pair. We need a representation. It is given by the predicted auc over all pairs of drugs. Then we use kmedoids on this representation.

Our goal is to design an initial batch of experiments for a new patient, under a cold-start assumption: no drug response data is available yet for this patient. In this setting, we seek to select a small number $K$ of drug-dose combinations that will be tested on patient-derived organoids, with the aim of quickly acquiring an informative profile of the patient’s drug sensitivity. This initial batch should maximize the downstream utility of the acquired data, enabling an optimal fine-tuning of the response model.

We decompose the experimental design process into two stages: selecting the drug pairs and selecting the dose pairs for each drug pair.

\subsubsection{Drug Pair Selection Strategy}

To select diverse and representative drug combinations, we rely on a learned representation of drug pairs. A natural choice for drug representations would be the pretrained embeddings \( e_d \in \mathbb{R}^p \). However, due to the presence of the multi-layer perceptron (MLP) in the model architecture, the spatial geometry of these embeddings is not straightforward to interpret: the MLP can effectively fold the embedding space, such that two drugs with distant embeddings \( e_{d_1} \) and \( e_{d_2} \) may ultimately yield similar response predictions. This nonlinearity complicates the use of \( e_d \) for measuring similarity or diversity between drugs.

Instead, we define each drug pair \( (d_1, d_2) \in \mathcal{D} \times \mathcal{D} \) by a vector of predicted area under the dose–response curve (AUC) across training samples \( s \in \mathcal{S} \), using the pretrained model \( f_\theta \). This yields a representation \( x \in \mathbb{R}^{|\mathcal{S}|} \), summarizing the pair's predicted effect across the population. This approach would not be feasible from raw data alone, since many drug pairs are sparsely tested; the model bridges these gaps with imputed predictions.

To select a diverse subset of drug pairs, we apply \( K \)-medoids clustering to the set \( \mathcal{X} = \{x_1, \ldots, x_n\} \subset \mathbb{R}^{|\mathcal{S}|} \) of drug-pair response vectors, with \( n = |\mathcal{D}|^2 \). The algorithm identifies a set of \( K \) medoids minimizing the total distance between each point and its closest medoid:
\[
\min_{\{m_1, \ldots, m_K\}} \sum_{i=1}^n \min_{k=1,\ldots,K} \|x_i - x_{m_k}\|_2.
\]
We use the FasterPAM algorithm \cite{schubert2019faster} for efficiency. For larger-scale settings, approximations such as OneBatchPAM \cite{deMathelin2025onebatchpam} or BanditPAM++ \cite{tiwari2023banditpam++} can be used.

Crucially, medoids are actual drug combinations, ensuring all selected items are experimentally testable. This selection provides broad coverage of the response landscape, as illustrated in Figure \ref{fig:kmedoids} in Section \ref{sec:results}.

\subsubsection{Dose Selection Strategy}
\label{dose-select-sec}

To select informative and diverse dose levels for each drug, we leverage the model’s predicted response curvature across training samples. Let \( K \) be the number of selected drug pairs, and \( n_d \) the number of times a drug \( d \in \mathcal{D} \) appears among them. Our goal is to choose \( n_d \) dose levels from the discretized range \( \{1, \ldots, L\} \) that reflect meaningful variations in the dose–response curve.

For each drug \( d \), we compute model-predicted dose–response curves over all training samples \( s \in \mathcal{S} \), and estimate the second derivative at each dose \( \ell \in \{1, \ldots, L\} \). To emphasize strong, sample-specific effects, we compute the 90th percentile at each dose, forming an importance score \( v_d \in \mathbb{R}^L \). 

Interpreting \( v_d \) as a density over the dose space, we select \( n_d \) doses that minimize the Wasserstein-1 distance between their empirical distribution and the distribution formed by \( v_d \), using a local search procedure. This yields dose sets focused on informative, high-curvature regions while maintaining good coverage. An illustration of the approach is provided in Figure \ref{fig:dose_select} in Section \ref{sec:results}.

This process is repeated for each drug, and the selected doses are randomly assigned to their corresponding pairs within the batch.

\section{Experiments}

\subsection{Dataset and Model}

For pretraining the model, we use the pan-preclinical (PPC) dataset \cite{markus2024pancancer}, a large-scale resource of drug response data across diverse patient-derived cells and cell lines. For evaluation, we use the GDSC$^2$ dataset \cite{jaaks2022effective}, a well-established pharmacogenomic screen of drug combinations across 126 cancer cell lines and 66 compounds. Currently, no ex vivo dataset of comparable scale exists for drug combinations, which motivates our choice of GDSC$^2$ for evaluation despite its reliance on in vitro models. To simulate a realistic cold-start scenario, all cell lines present in GDSC$^2$ are removed from the PPC dataset during training. This ensures that no response data from test cell lines is used during model pretraining.

% We evaluate our approach using the GDSC$^2$ \cite{jaaks2022effective} drug combination dataset, a large-scale pharmacogenomic screen that provides combination response data across 126 cancer cell lines and 66 compounds. This dataset spans several tumor types, including breast, colon, and pancreatic cancers, and offers detailed dose–response measurements for a diverse set of drug pairs.

% In our setup, we simulate a cold-start scenario in which no prior response is available for a new patient. To model this, we pretrain our drug response model on historical data (from other studies not including GDSC$^2$) to learn generalizable representations of drugs and doses. The GDSC$^2$ dataset is then used exclusively for evaluation, allowing us to assess how well the model performs in a new, unseen biological context.

While the GDSC$^2$ study provides valuable combination data, it does not contain exhaustive evaluations of all possible dose combinations for every drug pair. This introduces a bias toward the experimental design used in the original screen, which could unfairly favor active learning strategies that select combinations already present in the dataset. To mitigate this issue and evaluate our model uniformly across all drug–dose pairs, we train a “ground truth model” by fine-tuning the pretrained model on the full GDSC$^2$ dataset. We then use this fine-tuned model to generate predicted response values for all drug pairs across the full dose grid. These predicted values serve as standardized labels for benchmarking different selection strategies, ensuring an unbiased and comprehensive evaluation.

\subsection{Methodology}

To evaluate our cold-start active learning strategy, we follow a systematic experimental protocol that mirrors a real-world deployment scenario. Our methodology comprises model pretraining on a large dataset, simulating a new patient with no observed data, and selecting an informative batch of drug–dose combinations to guide fine-tuning. The full procedure is outlined below:

\begin{enumerate}
\item \textbf{Pretraining:} Train a prototype model on the PPC dataset, excluding all data from GDSC$^2$ and from cell lines present in GDSC$^2$. The MLP used for the model is composed of 5 hidden layers with 256 neurons each. The embedding dimension, $p$, is 128 and $r = 30$.
\item \textbf{Ground Truth Model:} fine-tune the pretrained model on the full GDSC$^2$ dataset to obtain a ground truth model. This model is then used to generate reference drug response values for all drug pair combinations. These imputed responses are used to simulate data acquisition and to evaluate the fine-tuned model without bias toward the specific experimental design of the GDSC$^2$ screen.

\item \textbf{Drug Representation:} Using the pretrained model, compute a representation vector for each drug pair by aggregating the AUCs of predicted responses across all training samples.

\item \textbf{Drug Pair Selection:} Perform K-medoids clustering on the drug pair representation vectors to select 100 representative drug pairs for experimentation.

\item \textbf{Dose Selection:} For each drug, select a number of doses equal to its number of occurrences in the selected drug pairs following procedure from Section \ref{dose-select-sec}.

\item \textbf{Assignment:} Assign the selected doses for each drug uniformly at random to the corresponding drug pairs in which the drug appears.

\item \textbf{fine-tuning:} Simulate data collection for a new sample by querying the ground truth model on the selected drug–dose combinations. Use the resulting responses to fine-tune the pretrained model for that sample.

\item \textbf{Evaluation:} Use the fine-tuned model to predict the viabilities of applying all possible drug/dose combinations present in the GDSC$^2$ dataset to the new sample. Compute the mean absolute error (MAE) of these predictions against the ground truth model predictions.
\end{enumerate}

\subsection{Baselines}

We compare our cold-start active learning strategy to several intuitive baselines. Importantly, conventional active learning approaches---which rely on estimating uncertainty or information gain using a representation of the test sample---are \textbf{not applicable in our setting}. This is because, in our cold-start scenario, we lack any observed response data for the new sample and therefore cannot construct a meaningful representation.

We consider the following baselines: \textbf{Random Drug Pairs + Random Doses:} Drug pairs are selected at random, and for each drug, a random set of doses is assigned. \textbf{Random Drug Pairs + Uniform Dose Grid:} Drug pairs are selected at random, and for each drug, the selected doses are chosen to be equally spaced across the dose axis. \textbf{Random Drug Pairs + Reweighted Dose Grid:} Drug pairs are selected at random, but dose selection follows our proposed strategy based on the second derivative of the drug response curves, prioritizing more informative dose regions. These baselines help isolate the contribution of both the drug pair selection strategy and the dose-level design in our method.

\subsection{Results}
\label{sec:results}

% Figure~\ref{fig:drug_response_surface} illustrates the effectiveness of our personalization strategy on a test cell line. The pretrained model shows systematic bias due to the cold-start setting, while the fine-tuned model, using only a small number of informative measurements, recovers a close approximation of the true response surface.

\begin{figure}[ht]
    \centering
    \includegraphics[width=\textwidth]{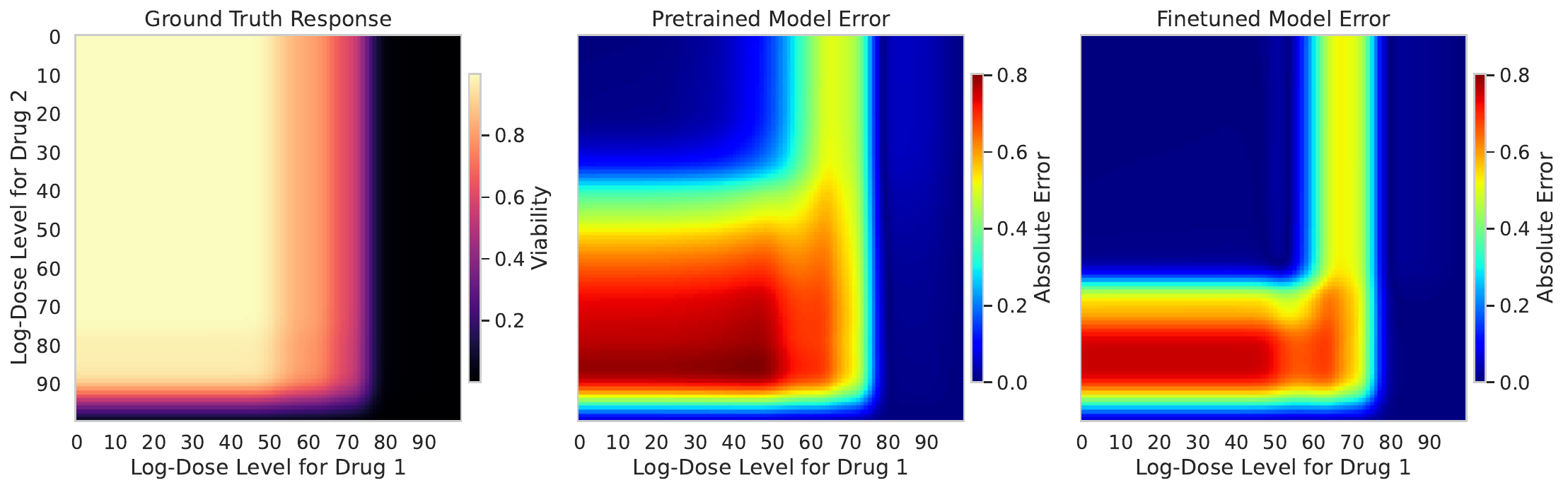}
    \caption{
        Predicted and true drug response surfaces for a given drug pair, for a held-out test sample from GDSC$^2$. The left panel shows the ground truth response over a $100 \times 100$ dose grid, imputed using a model fine-tuned on all observed viabilities from GDSC$^2$. The middle and right panels show the absolute prediction error of the pretrained and fine-tuned models, respectively. Because the pretrained model lacks an embedding for the test sample, its prediction corresponds to an average over training samples, reflecting a cold-start setting. The fine-tuned model is adapted to the test sample based on observed viabilities for the subset of drug–dose pairs selected by our cold-start approach.
    }
    \label{fig:drug_response_surface}
\end{figure}

We first show the comparison of the ground truth drug response and the predictions made by the pretrained and fine-tuned models for a test sample (Figure~\ref{fig:drug_response_surface}). Because the pretrained model has no embedding for the test sample, its output corresponds to the average drug response across all training samples. We observe that the fine-tuned model, informed by only a small set of selected measurements, significantly improves prediction accuracy and better captures the underlying response surface.

% In Figure~\ref{fig:kmedoids}, we visualize the drug pair selection process. K-medoids clustering yields a more diverse and representative coverage of the training distribution compared to random selection, which tends to oversample redundant or closely related drug pairs.

\begin{figure}[htbp]
    \centering
    \includegraphics[width=0.95\textwidth]{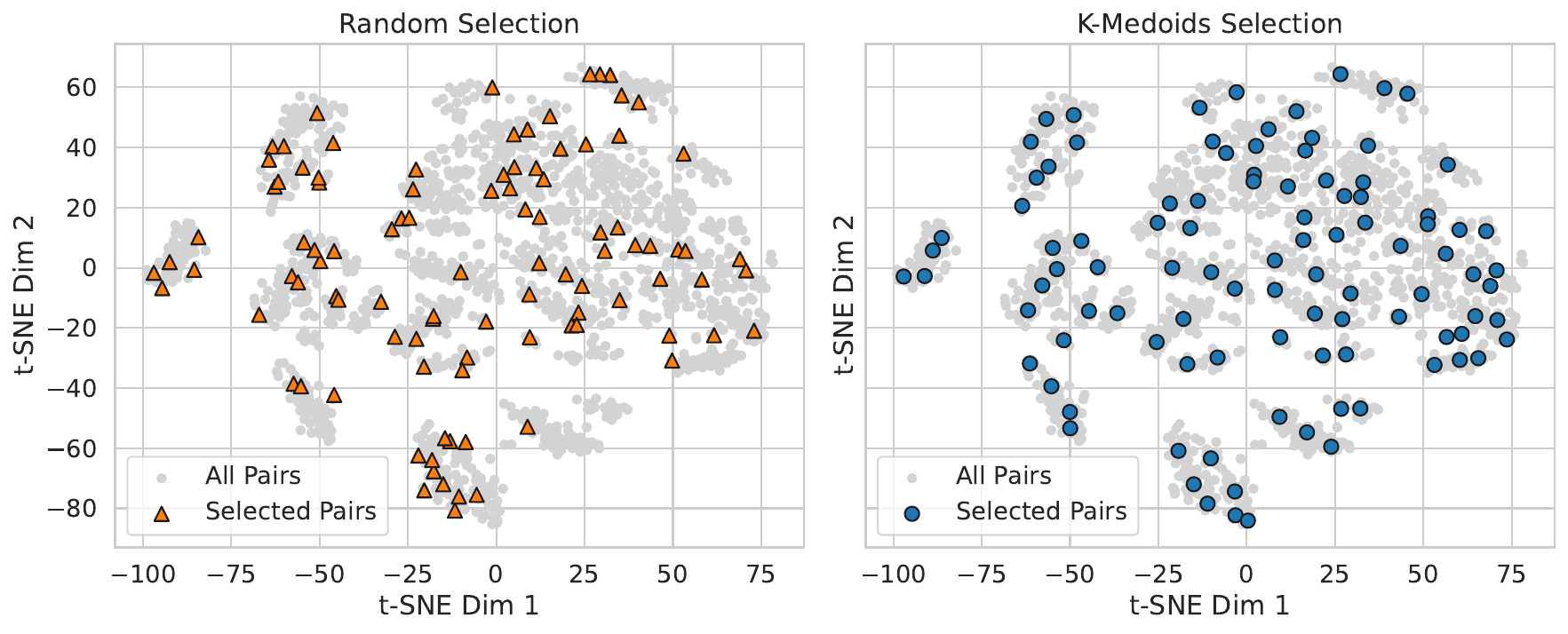}
    \caption{
        t-SNE visualization \cite{van2008visualizingtsne} of drug–pair response vectors, where each point represents a drug pair encoded by its vector of AUC values across training samples (in $\mathbb{R}^{|\mathcal{S}|}$, with $\mathcal{S}$ the set of training samples). Left: drug pairs selected at random. Right: pairs selected using K-medoids clustering, providing broader and more diverse coverage of the drug pairs.
    }
    \label{fig:kmedoids}
\end{figure}

\begin{figure}[htbp]
    \centering
    \includegraphics[width=\textwidth]{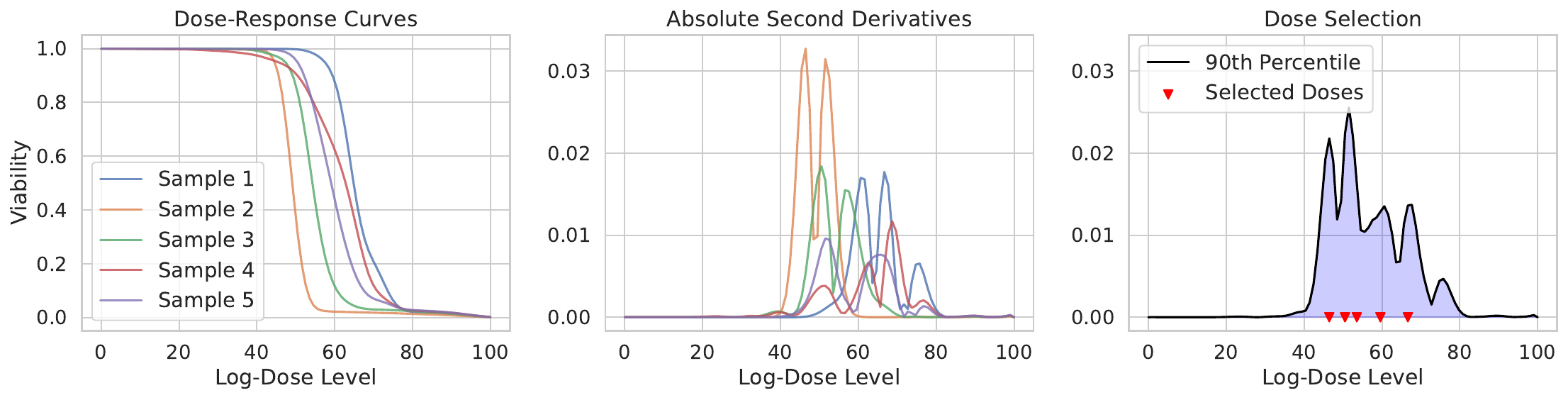}
    \caption{
        Dose selection procedure for a single drug in the set of selected drug pairs. We show dose-response curves (left) and their absolute second derivatives (middle) for one drug for 5 training samples from the training set $\mathcal{S}$. This drug appears in $5$ of the $100$ selected drug pairs. In the right panel, red triangles indicate 5 selected doses, chosen to minimize the Wasserstein-1 distance between their empirical distribution and the distribution given by the importance scores over doses, computed as the 90th percentile of absolute second derivatives across samples (shown in light blue).
    }
    \label{fig:dose_select}
\end{figure}

To illustrate the K-medoids selection strategy, we provide a comparison with random selection using a t-SNE visualization of the drug representation space (Figure~\ref{fig:kmedoids}). We observe that K-medoids selection provides a better coverage of the space, ensuring a more diverse and representative subset of drug pairs. Figure~\ref{fig:dose_select} illustrates our dose selection strategy for a single drug appearing in multiple selected pairs. The method selects doses that best match the distribution of importance scores, derived from curvature estimates of dose-response curves, leading to more informative measurements.

% Figure~\ref{fig:dose_select} highlights the dose selection mechanism. Our method selects doses that target regions of high curvature in the dose–response curve, maximizing information gain while maintaining an interpretable and principled allocation strategy.

\begin{figure}[htbp]
    \centering
    \includegraphics[width=0.95\textwidth]{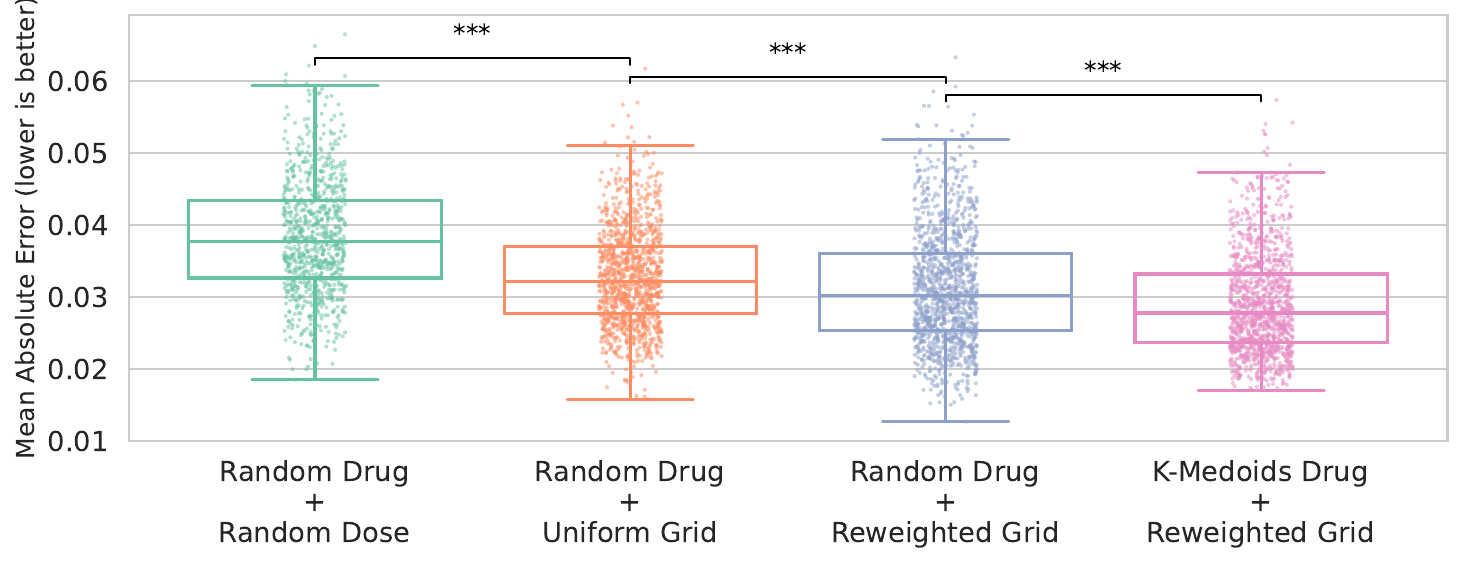}
    \caption{
        Comparison of acquisition strategies for cold-start personalization. Each method selects 100 drug–dose pairs used to fine-tune a predictive model for unseen samples. Boxplots show the distribution of predictive errors (MAE) across held-out samples.  Each dot represents the average MAE over 10 repetitions for one of the 126 held-out samples from GDSC\textsuperscript{2}. Horizontal black bars with asterisks indicate statistically significant improvements between adjacent methods (Wilcoxon signed-rank test, *** indicates $p < 0.001$). The performance improves as selection becomes more informed—first in dose selection, then in drug selection.
    }
    \label{fig:method_comparison}
\end{figure}

Finally, Figure~\ref{fig:method_comparison} compares predictive performance across acquisition strategies. Our full method achieves significantly lower error than all baselines, confirming the benefit of both informed drug pair and dose selection in cold-start scenarios.

\section{Conclusion and Future Directions}

We have presented a strategy to address the cold-start problem in personalized drug combination screening by leveraging pretrained drug response models to guide early experimental design. Our method combines clustering-based selection of informative drug pairs with a dose prioritization scheme that exploits curvature-based importance scores derived from historical data. This approach enables efficient personalization with limited experimental budget and no prior information about the test sample.

Once the cold-start phase is resolved and a small amount of data is collected, classical active learning strategies can be applied to further refine predictions. However, in ex vivo settings, the number of experimental rounds remains constrained due to limited tissue availability and resource demands, making the initial selection especially critical.

While our method offers substantial gains, several limitations and opportunities remain. In particular, our current dose-to-pair assignment strategy is arbitrary, and more principled allocation mechanisms could further improve performance. Additionally, our evaluation relies on cell lines data due to the lack of large-scale ex vivo combination screens. Future work will focus on applying our approach in real-world ex vivo experimental settings, which will enable prospective validation and may lead to more robust personalization pipelines for combination therapy.

% \bibliographystyle{plain}  % or use ieee, unsrt, apa, etc.
% \bibliography{sample}

%%%%%%%%%%%%%%%%%%%%%%%%%%%%%%%%%%%%%%%%%%%%%%%%%%%%%%%%%%%%

% \appendix

%%%%%%%%%%%%%%%%%%%%%%%%%%%%%%%%%%%%%%%%%%%%%%%%%%%%%%%%%%%%

\end{document}